\documentclass{article}
\usepackage{spconf,amsmath,graphicx}

\usepackage{booktabs}
\usepackage{hyperref}

\title{Objective, Absolute and Hue-aware Metrics for Intrinsic Image Decomposition on Real-World Scenes: A Proof of Concept}

\name{\em Shogo Sato, Masaru Tsuchida, Mariko Yamaguchi, Takuhiro Kaneko,\\ \em Kazuhiko Murasaki, Taiga Yoshida, Ryuichi Tanida\\}
\address{NTT Corporation, Kanagawa, Japan}

\begin{document}
%
\maketitle
\begin{abstract}
Intrinsic image decomposition (IID) is the task of separating an image into albedo and shade. In real-world scenes, it is difficult to quantitatively assess IID quality due to the unavailability of ground truth. The existing method provides the relative reflection intensities based on human-judged annotations. However, these annotations have challenges in subjectivity, relative evaluation, and hue non-assessment. To address these, we propose a concept of quantitative evaluation with a calculated albedo from a hyperspectral imaging and light detection and ranging (LiDAR) intensity. Additionally, we introduce an optional albedo densification approach based on spectral similarity. This paper conducted a concept verification in a laboratory environment, and suggested the feasibility of an objective, absolute, and hue-aware assessment.\footnote{Supplementary materials can be accessed \url{https://sigport.org/sites/default/files/docs/supp_Blinde_ID1165.pdf}}
\end{abstract}
\section{Introduction}
\label{sec:intro}
\begin{figure}
\centering
\includegraphics[width=0.9\linewidth]{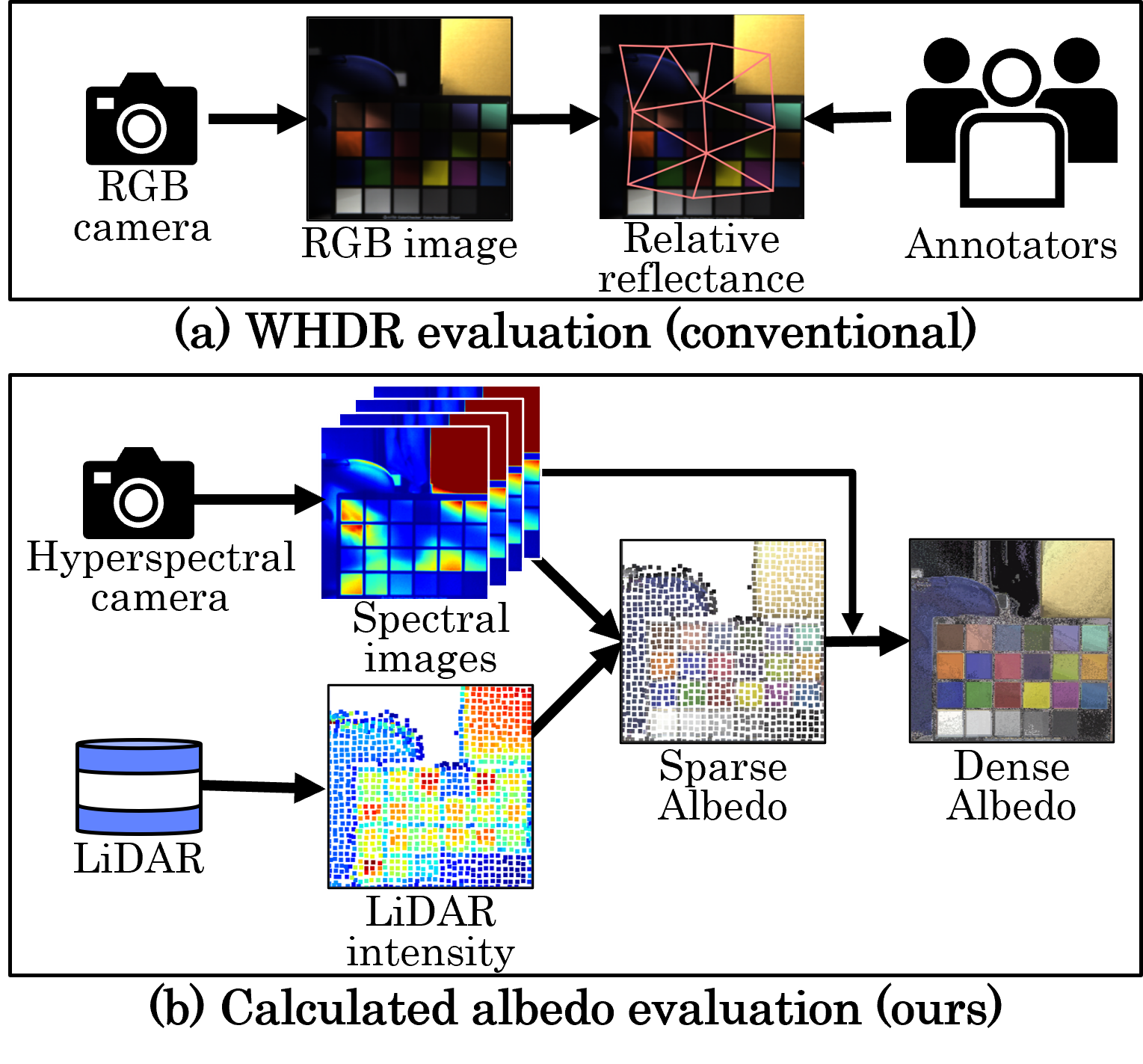}
\vspace{-5mm}
\caption{\label{fig1} (a) The existing method relies on human-judged annotations of relative reflectance, suffering from subjectivity, relative evaluation, and hue non-assessment. (b) our method evaluates IID quality with calculated albedo from hyperspectral images and LiDAR intensity, achieving objectivity, absolute-value evaluation, and hue-awareness. Optionally, the calculated albedo can be densified by spectral similarity algorithms.}
\vspace{-4mm}
\end{figure}

Intrinsic image decomposition (IID)~\cite{land1971} is the task of separating an image into illumination-invariant components (albedo) and illumination-variant components (shade). IID supports various high-level vision tasks such as semantic segmentation~\cite{wang2017robust}. IID methods are typically divided into two categories: synthetic data decomposition~\cite{li2018cg, wu2022uretinex} and real-world scene decomposition~\cite{fan2018, luo2020}. Synthetic data are generated through computer graphics and physics-based rendering~\cite{chang2015, li2018cg}, where known materials and light sources enable direct computation of ground-truth albedo and shade. Thus, the IID quality is simply evaluated with the ground truth. In contrast, evaluating IID on real-world scenes presents significant challenges in obtaining ground truths due to the difficulty of eliminating the effects of illumination. A common approach to address this is weighted human disagreement rate (WHDR) annotation~\cite{bell2014}, which collects human judgments on relative reflectance between point pairs in an image, as shown in Fig.~\ref{fig1} (a). Although WHDR has become the standard for IID evaluation in real-world scenes, it has challenges in subjectivity, relative evaluation, and hue non-assessment.

To address these issues, we propose a concept of IID evaluation in real-world scenes with calculated albedo from a hyperspectral camera and light detection and ranging (LiDAR) as illustrated in Fig.~\ref{fig1} (b). 
LiDAR captures reflectance at LiDAR wavelengths (LiDAR intensity) as well as depth. The hyperspectral camera captures images across a broad wavelength range, including the LiDAR wavelength. Reflectance at each wavelength is calculated from the ratio of LiDAR intensity to corresponding hyperspectral images. The reflectance spectrum is then converted into the RGB color space to compute albedo.

Our method does not require manual annotations of relative reflectance, offering objective, absolute-value evaluations. It also enables quantitative hue assessment through wavelength-based reflectance calculations. While the sparsity of LiDAR data may limit the density of calculated albedo, we propose a densification technique based on spectral similarity for applications requiring dense evaluations, such as model training. To verify the concept, we conduct an experiment on a color board with known albedo values in a laboratory environment. The main contributions of this study are summarized as follows:
\begin{itemize}
\item We propose a concept of IID evaluation in real-world scenes that computes albedo from hyperspectral images and LiDAR. Our method provides objective, absolute-value, and hue-aware evaluations.
\item Our method achieves high albedo quality with a CIEDE 2000 error of 6.75 on a reference color board, where an error of 7.0 indicates a 50\% acceptance rate. The luminance values show a correlation coefficient of 0.981 with ground truth, further validating the accuracy.
\item We introduce an albedo densification method based on spectral similarity to address sparsity, for applications requiring dense evaluations such as model training.
\end{itemize}

\section{Methodology}\label{sec3}
This section firstly reviews LiDAR intensity and hyperspectral images. Subsequently, we describe an albedo calculation process to evaluate IID quality. Then, our albedo densification method based on spectral similarity is described.

\subsection{\label{sec3-1}LiDAR intensity}
LiDAR aims to measure depth from time lags between laser emission and detection. In addition to depth, LiDAR also captures reflectance intensity, which indicates the proportion of emitted laser and reflected signal intensities~\cite{kashani2015review}. The LiDAR intensity $L$ can be modeled as Eq.~\ref{e1}.
\begin{equation}\label{e1}
L = \frac{D_r^2 \eta_{\text{sys}} \eta_{\text{atm}}}{4R^2} \rho(\lambda_{\text{LiDAR}}) \cos \theta,
\end{equation}
where $D_r$, $\eta_{\text{sys}}$, and $\eta_{\text{atm}}$ represent the receiver aperture diameter, system transmission factor, and atmospheric attenuation, respectively, which are determined by measurement conditions. Depth $R$ is obtained from the time lag between laser emission and detection, the incidence angle $\theta$ is derived from the depth map. Thus, reflectance at the LiDAR wavelength $\rho(\lambda_{\text{LiDAR}})$ can be calculated from LiDAR measurement. 

\begin{figure}
\centering
\includegraphics[width=1.0\linewidth]{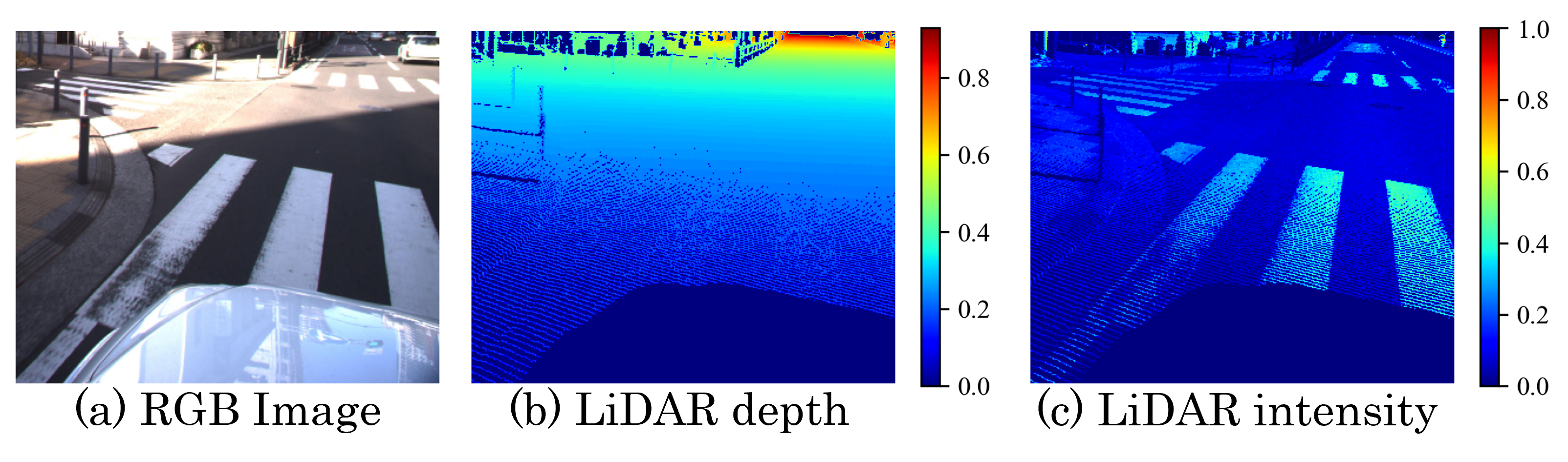}
\vspace{-8mm}
\caption{\label{fig2} (a) RGB image contains white line and cast shadows due to passive sensing. (b) LiDAR depth map dose not contain the white line and cast shadows. On the other hand, (c) LiDAR intensity contains white line while eliminating the cast shadows due to its active sensing.}
\vspace{-5mm}
\end{figure}

As illustrated in Fig.~\ref{fig2}, LiDAR intensity provides surface characteristics such as material reflectivity, independent of external lighting conditions like sunlight or shadows, contributing to the IID tasks~\cite{Sato2023, sato2025unsupervised}. However, due to the scanning process, LiDAR measurements are inherently sparse. This necessitates a densification step for applications requiring dense maps.

\subsection{Hyperspectral cameras}\label{sec3-2}
Hyperspectral cameras capture images across a broad wavelength range, providing detailed spectral information for each pixel. This spectral data enables precise material identification and analysis based on spectral signatures. To obtain the target image $I(\lambda)$ at a specific wavelength $\lambda$, the incident light spectrum $e(\lambda)$ is typically measured by capturing a whiteboard prior to scene acquisition.

\subsection{\label{sec4-1}Albedo calculation}
In IID setting, Lambertian surfaces are assumed, where only diffuse reflectance is considered. The hyperspectral pixel intensity $I(\lambda)$ at wavelength $\lambda$ can be described by Eq.~\ref{e3}.
\begin{equation}\label{e3}
I(\lambda) = m(\boldsymbol{n}, \boldsymbol{l}) e(\lambda) \rho(\lambda),
\end{equation}
where $m(\boldsymbol{n}, \boldsymbol{l})$ is a geometric factor depending on surface normal $\boldsymbol{n}$ and lighting direction $\boldsymbol{l}$, $e(\lambda)$ is the incident light intensity, and $\rho(\lambda)$ is the surface reflectance. Substituting $\lambda = \lambda$ and $\lambda = \lambda_{\text{LiDAR}}$ into Eq.~\ref{e3}, and then dividing the two expressions cancels out the common geometric factor:
\begin{equation}\label{e4}
\frac{I(\lambda)}{I(\lambda_{\text{LiDAR}})}
=\frac{e(\lambda) \rho(\lambda)}
{e(\lambda_{\text{LiDAR}}) \rho(\lambda_{\text{LiDAR}})}.
\end{equation}
This ratio expresses how the spectral reflectance at $\lambda$ compares to that at $\lambda_{\text{LiDAR}}$, independent of surface geometry. Rearranging this equation allows the reflectance $\rho(\lambda)$ to be computed as Eq.~\ref{e5}.
\begin{equation}\label{e5}
\rho(\lambda) = \frac{e(\lambda_{\text{LiDAR}})}{e(\lambda)} \frac{I(\lambda)}{I(\lambda_{\text{LiDAR}})} \rho(\lambda_{\text{LiDAR}}).
\end{equation}
Here, the illumination spectra $e(\lambda)$ and $e(\lambda_{\text{LiDAR}})$ are calibrated using a white reference target imaged prior to data acquisition. The intensity values $I(\lambda)$ and $I(\lambda_{\text{LiDAR}})$ are obtained from the hyperspectral image, and the LiDAR-based reflectance $\rho(\lambda_{\text{LiDAR}})$ is derived using Eq.~\ref{e1}. By spatially aligning the LiDAR data with hyperspectral imagery, we can recover the reflectance spectrum $\rho(\lambda)$ for each pixel. This spectrum is then transformed into the RGB color space using XYZ color matching~\cite{magnusson2020creating, wyszecki2000color}.

\begin{figure}
\centering
\includegraphics[width=0.9\linewidth]{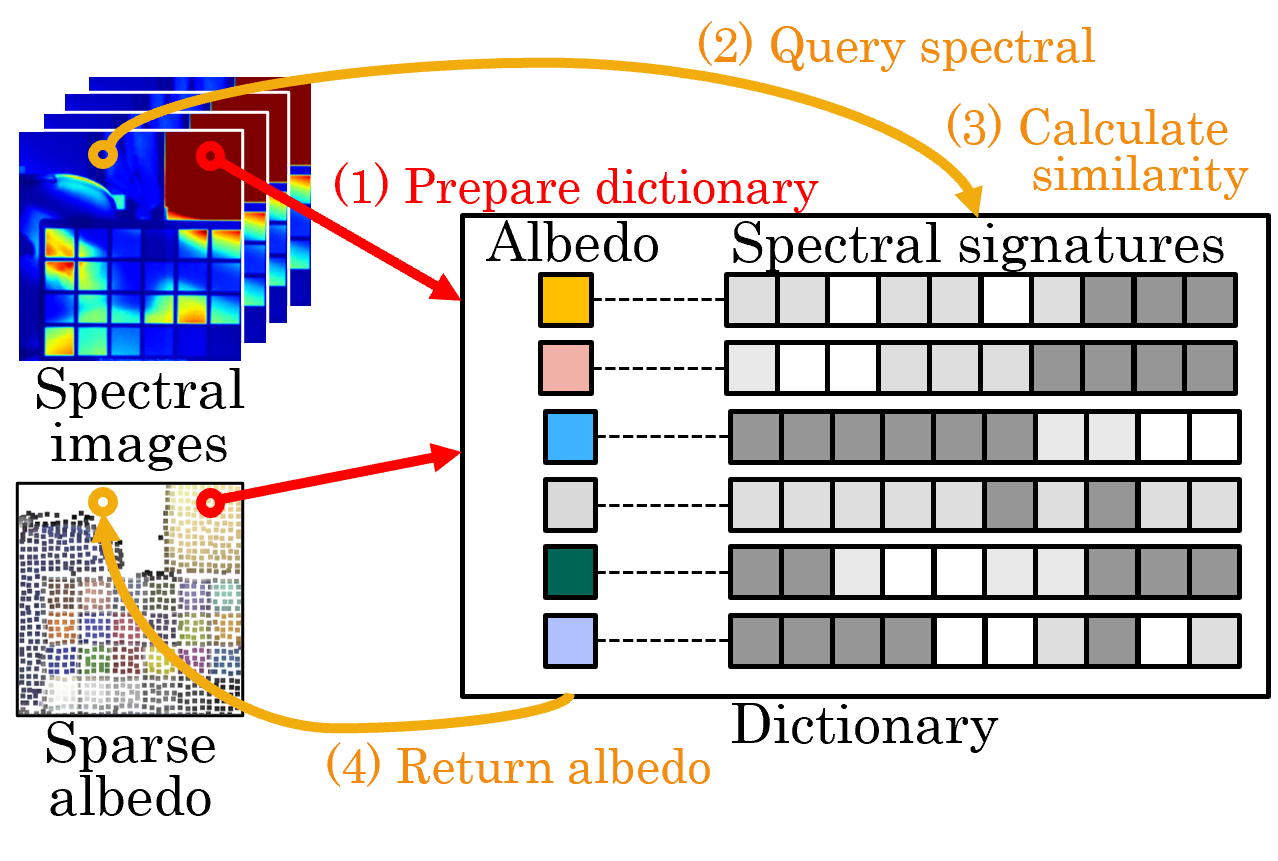}
\vspace{-5mm}
\caption{\label{fig3} Illustration of the proposed albedo densification method based on hyperspectral images. Step (1) is pre-processing to extract pixels with albedo values and prepare dictionary. Step (2) to (4) constitute the densification process: (2) use spectral data without albedo as query, (3) calculate similarity, and (4) assign densified albedo values. The spectral signatures represent that of each pixel.}
\vspace{-5mm}
\end{figure}

\subsection{\label{sec4-2}Albedo densification}
This section describes our albedo densification method, which leverages spectral similarity from hyperspectral images. Since LiDAR intensity-based albedo values are spatially sparse due to occlusions and sensor principles, dense estimation is required for image-wide reflectance reconstruction. Unlike typical inpainting or interpolation problems, this scenario benefits from physics-based priors, as hyperspectral observations contain rich spectral cues correlated with surface materials. The proposed procedure is illustrated in Fig.~\ref{fig3} and detailed below.

\noindent
1. \textbf{Dictionary construction from known albedo pixels.} For each pixel $(x_i, y_i)$ belonging to $P_1$ (the set of pixels with valid albedo measurements), we extract its spectral signature $f(x_i, y_i)$ from the hyperspectral image and store it in a dictionary, along with its albedo value $A(x_i, y_i)$. Each dictionary entry corresponds to a surface patch with known physical reflectance.

\noindent
2. \textbf{Querying unknown pixels.} For each pixel $(x_j, y_j)$ in $P_2$ (the set of pixels without albedo measurements), we extract its spectral signature $f(x_j, y_j)$. This signature is treated as a query to retrieve the most spectrally similar entries from the dictionary built in Step 1.

\noindent
3. \textbf{Spectral similarity search.} We compute a hybrid distance metric between the query signature $f(x_j, y_j)$ and each dictionary signature $f(x_i, y_i)$. The combined similarity incorporates both Euclidean and cosine distances:
\begin{equation}
\begin{aligned}
 \hat{i} = \arg\min_{i} &\Big[ \|f(x_j, y_j)-f(x_i, y_i)\|\\
&-\alpha \cos(f(x_j, y_j),f(x_i, y_i)) \Big],
\end{aligned}
\end{equation}
where $\cos(a, b)$ denotes the cosine similarity, and $\alpha$ is a weight parameter balancing intensity and angular similarity. The Euclidean term distinguishes intensity variations (e.g., between black and white surfaces), while the cosine term captures material-type similarity independent of shading.

\noindent
4. \textbf{Albedo assignment by neighborhood averaging.} To ensure robustness, the albedo value $\hat{A}(x_j, y_j)$ is assigned as the average of the albedo values corresponding to the three most similar dictionary entries:
\begin{equation}
\hat{A}(x_j, y_j) = \frac{1}{3} \sum_{k=1}^3 A(x_{\hat{i}_k}, y_{\hat{i}_k}).
\end{equation}
This averaging helps reduce quantization and noise effects caused by sparse dictionary entries and minor spectral distortions.

This non-parametric densification process is repeated for all pixels in $P_2$, yielding a spatially dense albedo map. The use of hyperspectral signatures as priors allows for semantically consistent interpolation, particularly effective in regions with material repetition or weak geometric texture.

\begin{figure}
\centering
\includegraphics[width=0.5\linewidth]{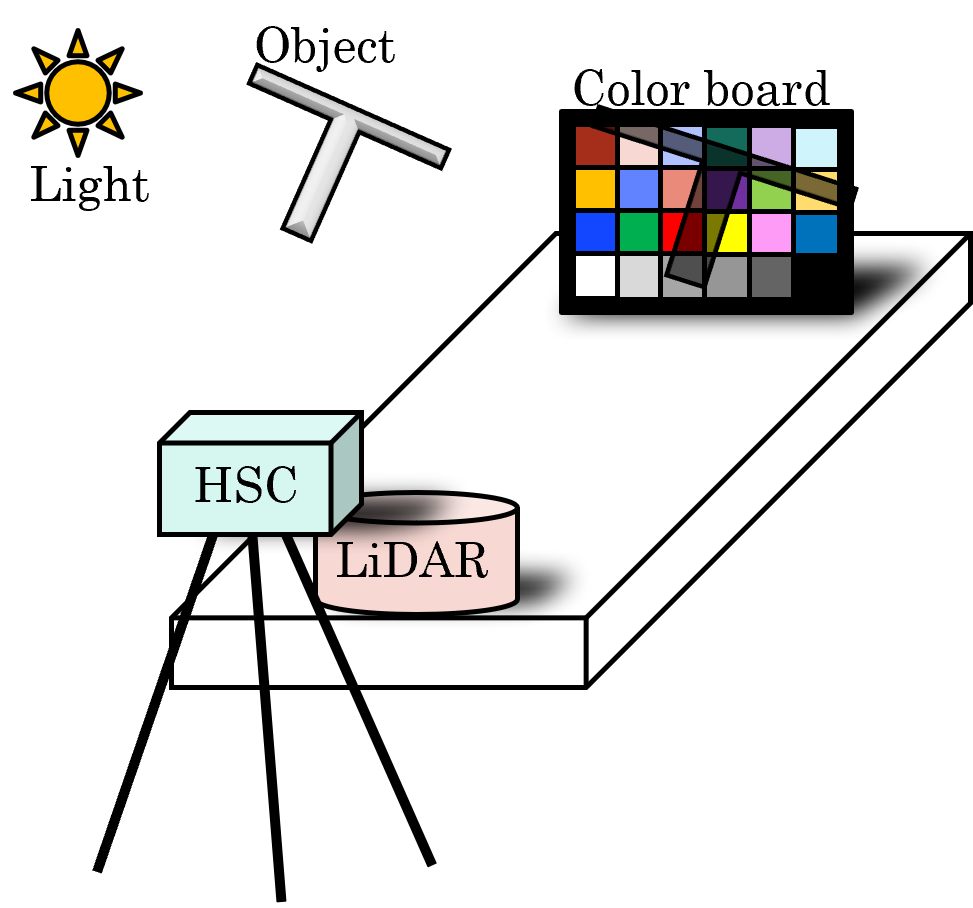}
\vspace{-5mm}
\caption{\label{fig4}Experimental setup. Hyperspectral camera, LiDAR, and color board are prepared to obtain experimental data. To illuminate the board, we use an artificial light source. Additionally, a T-shaped object is placed between the light source and the color board to create cast shadows.}
\vspace{-5mm}
\end{figure}

\section{\label{sec5}Experiments}
\subsection{\label{sec5-1}Experimental setting}
As a proof of concept, we conducted experiments using a hyperspectral camera, LiDAR, a color board, and an artificial illumination, as illustrated in Fig.~\ref{fig4}. As the hyperspectral camera, we used the SPECIM IQ, which covers a broad wavelength range, including the LiDAR wavelength. For LiDAR, we employed the Velodyne Alpha Prime, which provides high-density measurements. To quantitatively evaluate albedo calculation performance, we used the COLOR CHECKER CLASSIC from Calibrite, a standard reference for comparing calculated albedo values against known targets. Illumination was provided by a 500W artificial solar lamp from Solax, which replicates the solar spectrum to simulate outdoor conditions. The hyperspectral and LiDAR data were processed to calculate albedo as described in Sec.~2. The weight parameter $\alpha$ for spectral similarity was empirically set to 1.0, as noted in supplementary materials.

\begin{figure}
\centering
\includegraphics[width=1.0\linewidth]{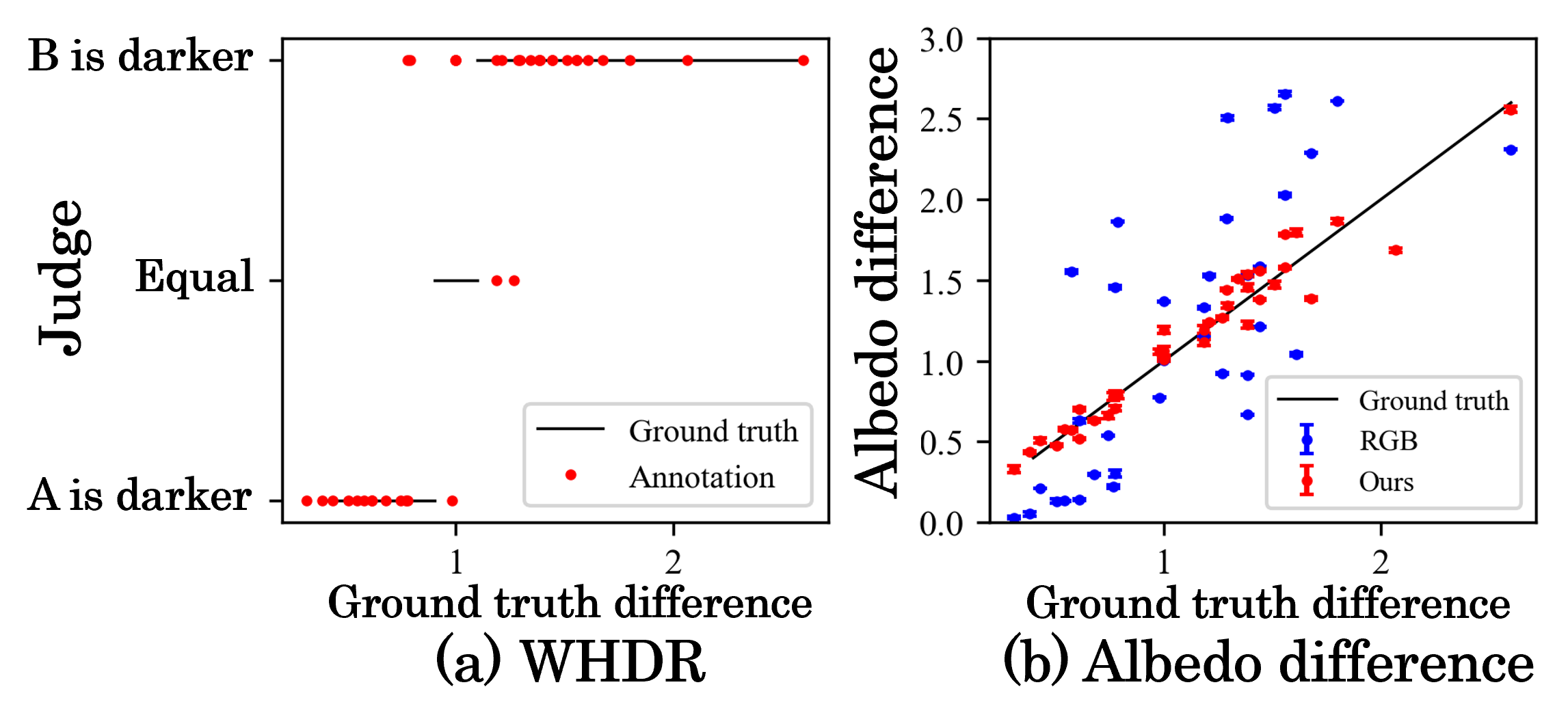}
\vspace{-8mm}
\caption{\label{fig8} Comparison between (a) WHDR annotation and (b) our proposed evaluation, by plotting color differences within the color board image. WHDR relies on human-judged annotations, leading to subjectivity and discrete-value evaluation. In contrast, ours computes albedo from hyperspectral images and LiDAR intensity, providing objective and continuous-value evaluation. Furthermore, compared to the RGB image, our albedo plots align closely with the ground truth, forming an almost straight line.}
\end{figure}

\subsection{Comparison with WHDR annotations}\label{sec5-6}
WHDR~\cite{bell2014} is a widely used metric for IID evaluation in real-world scenes, relying on human-judged annotations to compare relative reflectance. However, it has limitations, including subjectivity, relative evaluation, and hue non-assessment. 

To compare our method with WHDR, we annotated the color board area in Fig.~\ref{fig5} (a) following the existing paper~\cite{bell2014}. Using the luminance values of point A ($L_{\rm{A}}^{\rm{G}}$) and point B ($L_{\rm{B}}^{\rm{G}}$), we calculated the ground-truth difference ($L_{\rm{A}}^{\rm{G}}/L_{\rm{B}}^{\rm{G}}$). Fig.~\ref{fig8} (a) illustrates the relationship between the ground-truth luminance differences and the WHDR annotations. Ideally, when the difference is close to 1, the annotation should be "Equal". When the difference is small, indicating point A is darker, the annotation should be "A is darker", and vice versa for point B. Consequently, there were errors in 8 of 38 annotations, especially in areas darkened by cast shadows. 

Fig.~\ref{fig8} (b) compares the albedo difference calculated by our method with the ground-truth. The scatter plot shows the correlation between our method (red) and RGB values from hyperspectral images (blue). The black line represents the ideal case where predicted differences perfectly match ground truth. In the RGB plots, many points deviate from the ideal line due to cast shadows. In contrast, most points from our method align closely with the ideal line, accurately reflecting luminance values. This demonstrates that our method provides objective and absolute-value evaluations, while WHDR is subjective and relies on relative-value evaluations.

\tabcolsep = 1pt
\begin{table}
\centering
\small
\begin{tabular}{@{}ccccccccc@{}}
\toprule
Learning & Method & CIE76 &\ & CIEDE &\ & Coeffs &\ & MSE($\times10^{-3}$) \\
\midrule
No & RGB image                       & 37.8 && 26.8 &&0.807&&94.3\\
No & Retinex~\cite{grosse2009}       & 33.2 && 23.7 &&0.889&&70.1\\
No & Bell et al.~\cite{bell2014}     & 22.8 && 14.9 &&0.808&&39.1\\
No & Ours (Sparse)& \textbf{13.5} && \textbf{6.75}  &&\textbf{0.981}&&\textbf{7.83}\\
No & Ours (Dense) & \textbf{13.4} && \textbf{6.73}  &&\textbf{0.978}&&\textbf{8.04}\\
\midrule
Sup. & Revisit~\cite{fan2018}        & 30.9 && 20.1 &&0.747&&49.3\\
Unsup. & IIDWW~\cite{li2018}         & 24.0 && 16.1 &&0.806&&22.7\\
Unsup. & IID-LI~\cite{Sato2023}      & 42.3 && 28.2 &&0.393&&103\\
\bottomrule
\end{tabular}
\caption{\label{tab1}
Quantitative evaluation of the color board area based on CIE76, CIEDE2000, luminance correlation coefficients (Co-effs), and MSE. The RGB image was derived from hyperspectral data. Existing rule-based IID methods including Retinex~\cite{grosse2009}, and Bell et al.~\cite{bell2014} are evaluated. In addition, supervised model (sup.): Revisit~\cite{fan2018}, and unsupervised models (unsup.): IIDWW~\cite{li2018}, and IID-LI~\cite{Sato2023} are assessed. As reported in a previous study~\cite{biswas2022does}, the acceptance rate is 50\% when the value is about 7.0 in CIEDE 2000.}
\end{table}

\subsection{Albedo calculation quality}\label{sec5-4}
In the previous section, we demonstrated the advantages of using our calculated albedo for quantitative evaluation over WHDR, highlighting its objectivity and absolute-value assessment capabilities. However, a potential critique is why albedo inferred from existing IID models cannot be used for similar evaluations. To address this concern, we compare the albedo estimation performance of our method with that of traditional IID models in this section. The compared models includes rule-based methods like Retinex~\cite{grosse2009}, and Bell et al.~\cite{bell2014}, as well as deep learning-based models such as Revisit~\cite{fan2018}, IIDWW~\cite{li2018}, and IID-LI~\cite{Sato2023}. For these comparisons, we used publicly available pre-trained parameters. 

To evaluate our proposed method, we used a color board with known reference colors. Quantitative metrics included CIEDE2000 and CIE76 color difference formulas, mean squared error (MSE), and luminance correlation coefficients to assess grayscale quality. Tab.~\ref{tab1} shows that our method achieves a CIEDE2000 score of 6.75 and a CIE76 score of 13.5, significantly outperforming the RGB image (26.8 and 37.8) and other IID methods such as Bell et al.~\cite{bell2014} (14.9 and 22.8). The luminance correlation coefficient with ground truth was 0.981 for our method, compared to 0.807 for the RGB image, highlighting its robustness against cast shadows.

Visually, Fig.~\ref{fig5} demonstrates the effectiveness of our method. Fig.~\ref{fig5} (a) is ground truth image created by manually mapping the correct color on the color board. The RGB image derived from hyperspectral data in Fig.~\ref{fig5} (b) contains noticeable cast shadows. While conventional methods, such as Bell et al.~\cite{bell2014}, provide reasonable results, the shadowed regions remain darker. IID-LI, which uses RGB values and LiDAR intensity, was particularly sensitive to data domain gaps, leading to challenges in estimation without training data. In contrast, our method accurately reconstructs colors even in shadowed areas, providing objective and robust albedo estimation. Sources of noise and limitations of our approach are further discussed in Sec.~\ref{sec5-7}.

\subsection{\label{sec5-5}Albedo densification quality}
This section describes the evaluation of albedo densification. Due to the absence of high-density ground truth, we evaluated it through visual inspection. Our method was compared with conventional approaches, including nearest-neighbor completion~\cite{telea2004}, DIP~\cite{ulyanov2018}, and DIP with RGB prior~\cite{Sato2023}, none of which require training data.

\begin{figure}
\centering
\includegraphics[width=1.0\linewidth]{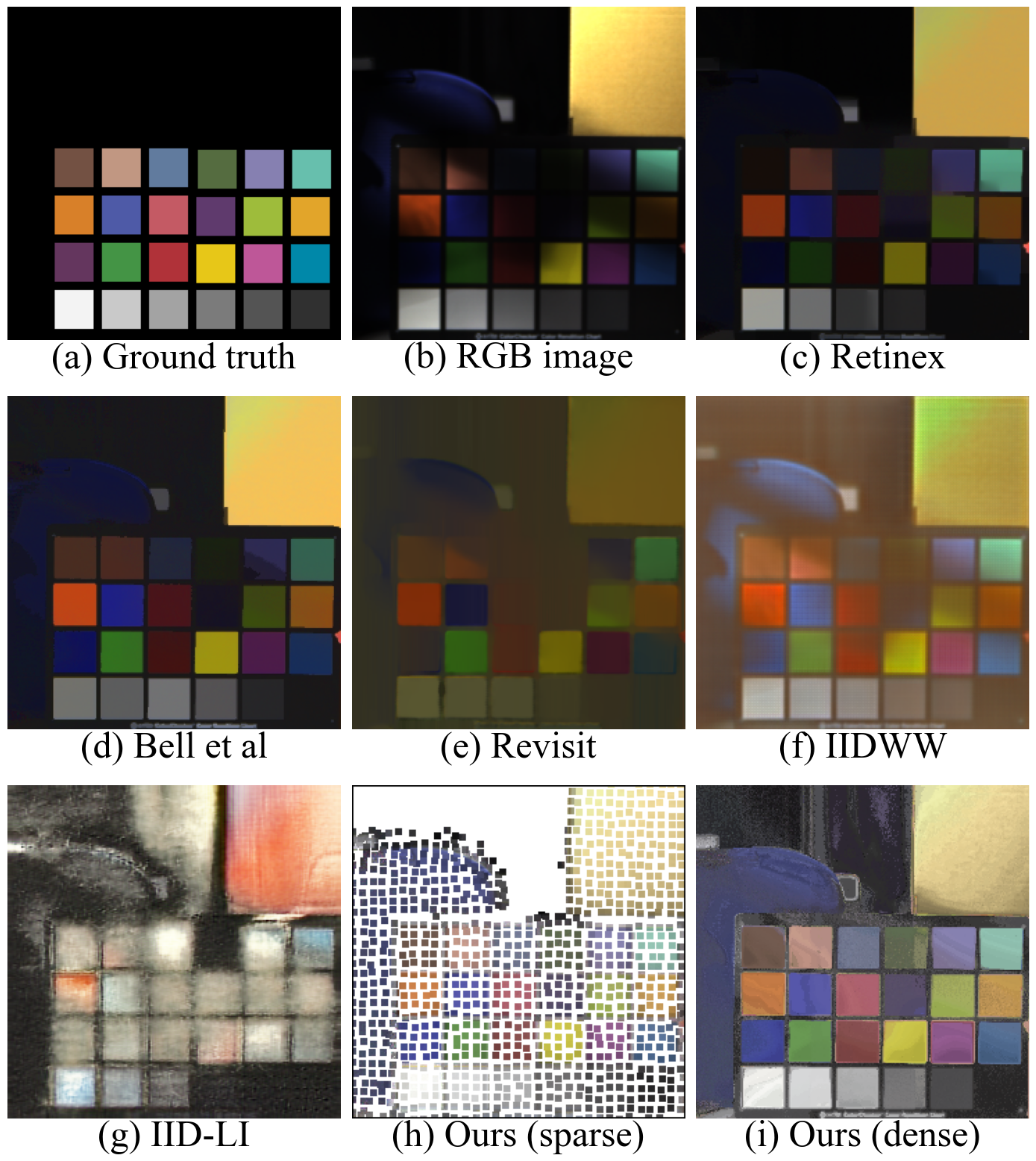}
\vspace{-8mm}
\caption{\label{fig5} Visual results of calculated albedo. We depict (a) ground-truth color, (b) original RGB image, (c) Retinex~\cite{grosse2009}, (d) Bell et al.\cite{bell2014}, (e) Revisit\cite{fan2018}, (f) IIDWW~\cite{li2018}, and (g) IID-LI~\cite{Sato2023}. Our calculated albedos with (h) sparse and (i) dense are also shown. Although shadows are present in the RGB image and existing methods, ours eliminates the shadows.}
\vspace{-4mm}
\end{figure}

\begin{figure}
\centering
\includegraphics[width=0.9\linewidth]{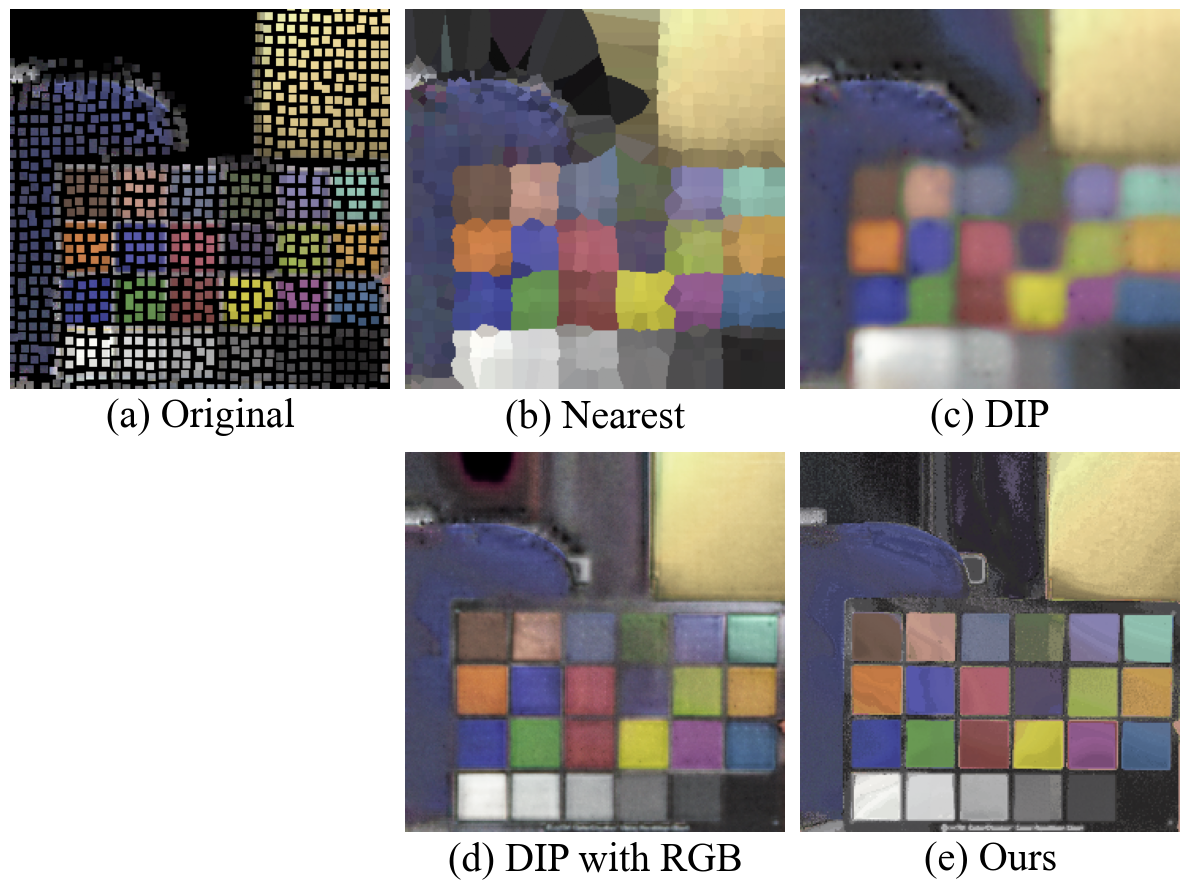}
\vspace{-4mm}
\caption{\label{fig6}Visual results of albedo densification. (a) Original albedo, (b) Albedo completed using nearest-neighbor values~\cite{telea2004}, (c) albedo densified using deep image prior, (d) albedo densified using RGB and sparse albedo with deep image prior, and (e) proposed method.}
\end{figure}

Fig.~\ref{fig6} illustrates the visual outcomes of albedo densification for the sparse albedo in Fig.~\ref{fig5}(d). In Fig.~\ref{fig6}(b), nearest-neighbor interpolation~\cite{telea2004} fills sparse regions but introduces artifacts and inconsistencies. Fig.~\ref{fig6}(c) and (d) shows results from the original DIP~\cite{ulyanov2018} and DIP with an RGB prior~\cite{Sato2023}, respectively. These models improves density but still retains some artifacts and lacks edge preservation. In contrast, our method in Fig.~\ref{fig6}(e) produces dense, edge-preserved albedo with minimal artifacts and consistent values across regions.

\subsection{Quality limitation factors}\label{sec5-7}
This section discusses the primary factors limiting the quality of our albedo calculation and densification method:

\noindent
1. \textbf{Precision of $\rho(\lambda_{\text{LiDAR}})$.} The accuracy of LiDAR intensity, as noted in Eq.~\ref{e1}, is dependent on the performance of surface-normal estimation. Since our algorithm derives normals from LiDAR depth, the precision of $\rho(\lambda_{\text{LiDAR}})$ is constrained by the density of LiDAR points. Thus, increasing the LiDAR density enhances the accuracy of $\rho(\lambda_{\text{LiDAR}})$.

\noindent
2. \textbf{Noise in hyperspectral images.} Hyperspectral image noise, primarily caused by electronic circuits, reduces the quality of dark regions where weak signals are overwhelmed by noise. As shown in Fig.~\ref{fig5}(a), noise affects albedo calculation and densification in extremely dark areas. However, in real-world scenarios, sunlight is 10–100 times brighter than artificial sunlamps, making circuit noise negligible.

\noindent
3. \textbf{Illuminated light distribution.} 
The proposed method calculates albedo using the ratio of incident light spectra \(e(\lambda_{\text{LiDAR}})/e(\lambda)\) for each pixel. When the incident light spectrum is approximately uniform across the scene, calculating this ratio for a representative pixel is sufficient. However, in cases where significant spatial variations exist in the incident light spectrum, it becomes necessary to measure \(e(\lambda)\) in multiple regions using a whiteboard. This supports accurate albedo calculation even in complex lighting environments.

By addressing these factors, our method can be expanded to more complex scenes in real world.

\section{Conclusion}\label{sec6}
This paper proposed the concept of albedo calculation for assessing real-world IID using hyperspectral images and LiDAR intensity. Our method objectively computes the absolute albedo values along with hue assessment. Achieving an error rate of approximately 6.75 (CIEDE 2000 standard), our proposal comfortably exceeds the acceptance threshold in a laboratory environment, suggesting its practical feasibility. For future work, we plan to expand the experimental setup to include more complex scenes.

\vfill\pagebreak
\bibliographystyle{IEEEbib}
\bibliography{refs}

\end{document}